\documentclass[sigconf, screen, nonacm]{acmart}

\usepackage{booktabs} 
\usepackage{enumitem} 
\usepackage{graphicx}
\usepackage{subcaption}
\usepackage{algorithm}
\usepackage{algpseudocode}
\usepackage[font=small,labelfont=bf]{caption}  

\usepackage{booktabs}
\usepackage{multirow}
\usepackage{siunitx}
\usepackage{hyperref}
\usepackage{hyperxmp}
\usepackage{wrapfig}
\usepackage{float}

\definecolor{myblue}{HTML}{3333FF}
\usepackage{tikz}
\newcommand{\bluecircled}[1]{%
  \begin{tikzpicture}[baseline=(char.base)]
    \node[shape=circle, draw=myblue, fill=myblue, text=white, inner sep=1pt] (char) {#1};
  \end{tikzpicture}%
}

\definecolor{myviolet}{HTML}{6600CC}
\newcommand{\violetcircled}[1]{%
  \begin{tikzpicture}[baseline=(char.base)]
    \node[shape=circle, draw=myviolet, fill=myviolet, text=white, inner sep=1pt] (char) {#1};
  \end{tikzpicture}%
}

\renewcommand\footnotetextcopyrightpermission[1]{}
\acmConference{}{}{}
\acmBooktitle{}
\acmYear{}
\acmDOI{}
\acmISBN{}
\begin{document}
\title{Unveiling Text in Challenging Stone Inscriptions: A Character-Context-Aware Patching Strategy for Binarization}

\author{Pratyush Jena, Amal Joseph, Arnav Sharma, Ravi Kiran Sarvadevabhatla}
\email{(pratyush.jena, amal.joseph, arnav.sharma)@research.iiit.ac.in, ravi.kiran@iiit.ac.in}
\affiliation{%
  \institution{Center for Visual Information Technology, International Institute of Information Technology, Hyderabad}
  \streetaddress{Gachibowli}
  \city{Hyderabad}
  \state{Telangana}
  \country{India}
  \postcode{500032}
}

\renewcommand{\shortauthors}{}

\begin{abstract}
Binarization is a popular first step towards text extraction in historical artifacts. Stone inscription images pose severe challenges for binarization due to poor contrast between etched characters and the stone background, non-uniform surface degradation, distracting artifacts, and highly variable text density and layouts. These conditions frequently cause existing binarization techniques to fail and struggle to isolate coherent character regions. Many approaches sub-divide the image into patches to improve text fragment resolution and improve binarization performance. With this in mind, we present a robust and adaptive patching strategy to binarize challenging Indic inscriptions. The patches from our approach are used to train an Attention U-Net for binarization.
The attention mechanism allows the model to focus on subtle structural cues, while our dynamic sampling and patch selection method ensures that the model learns to overcome surface noise and layout irregularities. We also introduce a carefully annotated, pixel-precise dataset of Indic stone inscriptions at the character-fragment level. We demonstrate that our novel patching mechanism significantly boosts binarization performance across classical and deep learning baselines. Despite training only on single script Indic dataset, our model exhibits strong zero-shot generalization to other Indic and non-indic scripts, highlighting its robustness and script-agnostic generalization capabilities. By producing clean, structured representations of inscription content, our method lays the foundation for downstream tasks such as script identification, OCR, and historical text analysis.
Project page: \url{https://ihdia.iiit.ac.in/shilalekhya-binarization/}
\end{abstract}

%
%


\maketitle
\section{Introduction}

Stone inscriptions are rich sources of historical and linguistic knowledge, but their automated analysis remains challenging. Unlike scanned documents or manuscripts, they often exhibit severe degradation—shallow etching, erosion, surface noise, and uneven lighting. Text layout varies widely, and inscriptions frequently include decorative or non-textual elements, making standard image processing unreliable.


Binarization is a key step for text detection and document understanding. Traditional methods like Otsu~\cite{otsu1979} and Sauvola~\cite{sauvola2000} often fail on degraded inscriptions with shallow or worn characters. Deep learning models such as U-Net~\cite{Ronneberger2015} offer improved performance but rely on annotated data and fine-grained spatial understanding. Patching is commonly used to handle high-resolution images and varying text densities, yet often treated as a trivial detail. We argue that patch design is critical: poorly chosen patches can lack context or visual representativeness, degrading predictions. A principled patching strategy can significantly improve performance at minimal cost.


In this work, we focus on pixel-precise, character-level binarization for challenging stone inscriptions. We introduce a pipeline that uses an Attention U-Net trained with a patching strategy tailored to the inscription dimensions and typical character heights. This ensures each patch contains sufficient context for the model to distinguish foreground characters from the noisy stone background and non-textual carvings. To support training and evaluation, we construct a high-quality dataset of 203 annotated stone inscriptions with character fragments labeled at fine granularity. Although trained solely on one Indic script, our model generalizes well to other Indic and non-Indic scripts, demonstrating robustness in zero-shot scenarios. Our method outperforms both traditional and learning-based baselines, and incorporating our patching strategy also improves baseline performance.

Beyond solving a difficult binarization task, our approach lays a strong foundation for downstream epigraphic analysis, including script identification, OCR, and semantic region interpretation.

\begin{figure*}[!t]
    \centering
    \includegraphics[width=0.9\textwidth]{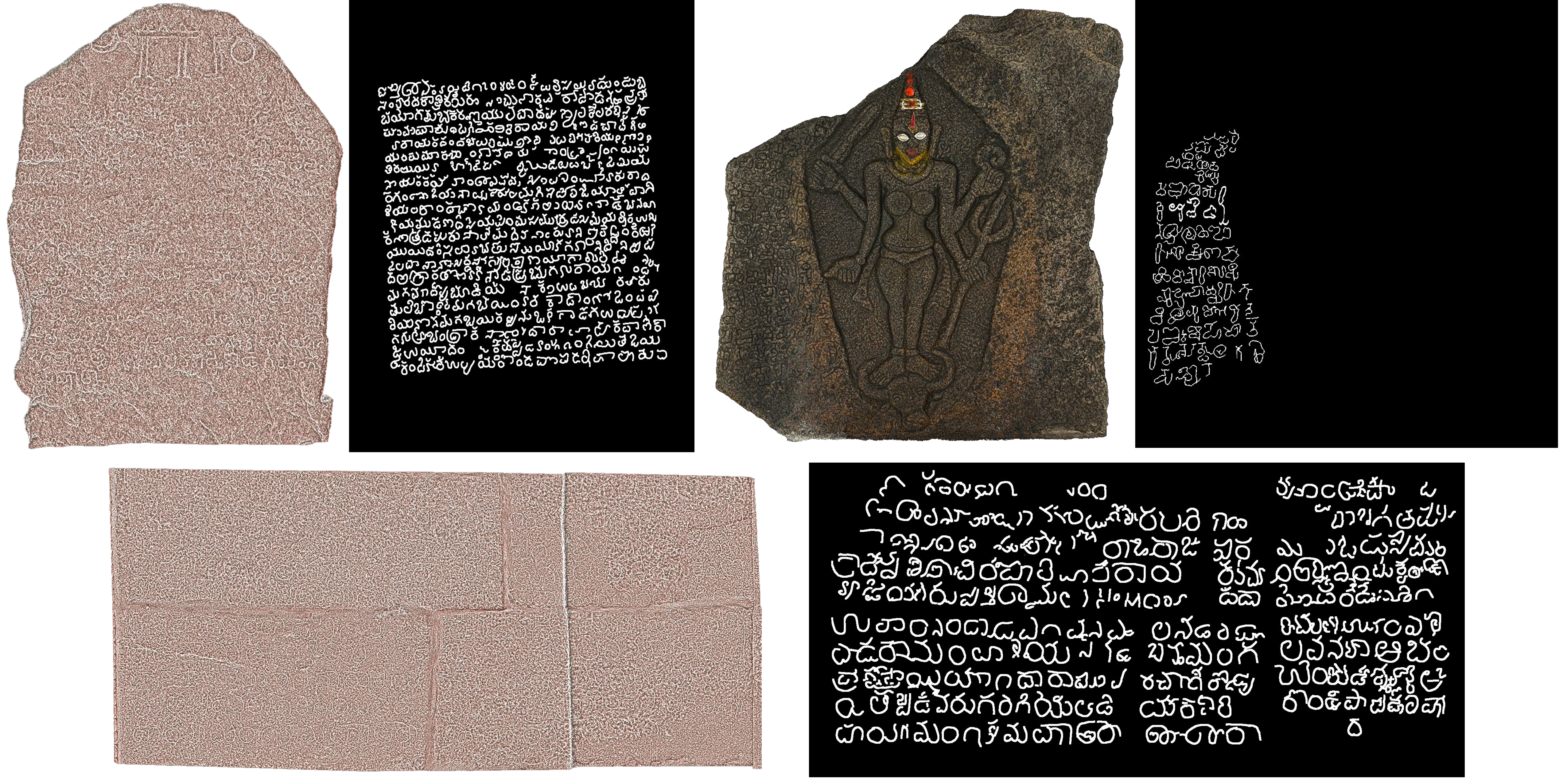}
    \caption{Sample images and their corresponding binarization ground truth from the annotated stone inscription dataset. Notice the difficulty distinguishing the shallow handwritten text etching from the background stone texture with naked eye.}
    \label{fig:dataset-samples}
\end{figure*}




\section{Related Work}

\noindent \textbf{Document Image Binarization:}
Classical methods like Otsu~\cite{otsu1979}, Niblack, and Sauvola~\cite{niblack1985introduction, sauvola2000} rely on global or local intensity thresholds. Though efficient, they struggle with degraded or textured backgrounds typical of stone inscriptions. Contrast-enhancement-based methods~\cite{Su2013}, and MRF-based formulations like Howe's~\cite{Howe2013} offer more robustness but still under perform on complex cases.

Deep learning methods have led to significant improvements in binarization. 
U-Net~\cite{Ronneberger2015} and its variants (including Attention U-Net~\cite{Oktay2018}), as well as Fully Convolutional Networks optimized for document binarization~\cite{Tensmeyer2017}, have shown strong performance in pixel-wise segmentation. The DIBCO 2019~\cite{8978205} leader board represents the current state of the art. NAF-DPM~\cite{cicchetti2024nafdpm} utilizes a diffusion probabilistic model while ~\cite{yang2024gdb} follows a gated convolutional architecture. ~\cite{biswas2023layer,souibgui2022docentr} follow the Transformer ~\cite{vaswani2017attention} architecture.

\noindent \textbf{Adaptive Patching Mechanisms:}
Patching is widely used in document analysis to handle high-resolution images and augment training data. Most approaches adopt a fixed patch size with overlap~\cite{souibgui2022docentr,cicchetti2024nafdpm,Vadlamudi2023}, which works well for uniform layouts. However, in stone inscriptions where character size and spacing vary significantly, fixed-size patching often fails—either missing entire characters or including excessive background, weakening the model’s focus.

Adaptive strategies have been explored, but are not well-suited for epigraphy. LineTR~\cite{agrawal2025linetr} adapts patch size based on interline gaps, which are inconsistent or absent in inscriptions. Quad-tree decomposition~\cite{8451252} leads to overly small patches with poor context, and spatially-adapted sliding windows~\cite{hedjam2011spatially,koloda2023context} struggle to maintain coherence across scales. These limitations motivate our custom patching strategy tailored to the structure and variability of stone inscriptions.

\noindent \textbf{Epigraphy and Stone Inscription Analysis:}  Computational work on stone inscriptions across different scripts remains limited. A recent overview of computational epigraphy surveys modern approaches including template matching and edge-based filtering~\cite{Vishal2024}. For Indian scripts, HOG+SVM has been used to recognize ancient Tamil inscription characters~\cite{Bhuvaneswari2019}, while CNN-based OCR for Ashokan Brahmi demonstrates strong performance using transfer learning~\cite{Agrawal2024}. Adhikari and Palaniappan~\cite{Palaniappan2017} proposed a deep-learning pipeline for segmenting and classifying symbols in Indus script seal impressions. In Greek and Roman studies, image enhancement and template matching have been applied to recover faint carvings from weathered surfaces~\cite{Alvarez2010}. Munivel et al.~\cite{munivel2024mlibt} propose a multi-level binarization technique for Tamil inscriptions, but report limited robustness under severe degradation.

End-to-end recognition systems often assume clean segmentation and high contrast, which is rarely the case in inscriptions. Erosion, poor lighting, and fragmented layouts degrade recognition performance. Our method instead focuses on robust binarization, forming a reliable, modular foundation for OCR and script analysis across scripts and degradation levels.

\section{Dataset}
We introduce a new dataset of 203 high-resolution images of stone inscriptions carved in the  script. The inscriptions span diverse historical periods, styles of etching, and physical conditions, including erosion, moss growth, cracks, and inconsistent lighting. These stone inscriptions are extremely challenging as the etching strokes and the background surface are visually indistinguishable from noise. Along with text, there are visual elements like reliefs and iconography which further increase the complexity. Each image has been carefully annotated at the character-fragment level, resulting in fine-grained binary masks for every distinguishable fragment (See Fig. ~\ref{fig:dataset-samples}).

\begin{figure*}[!t]
    \centering
    \includegraphics[width=\textwidth]{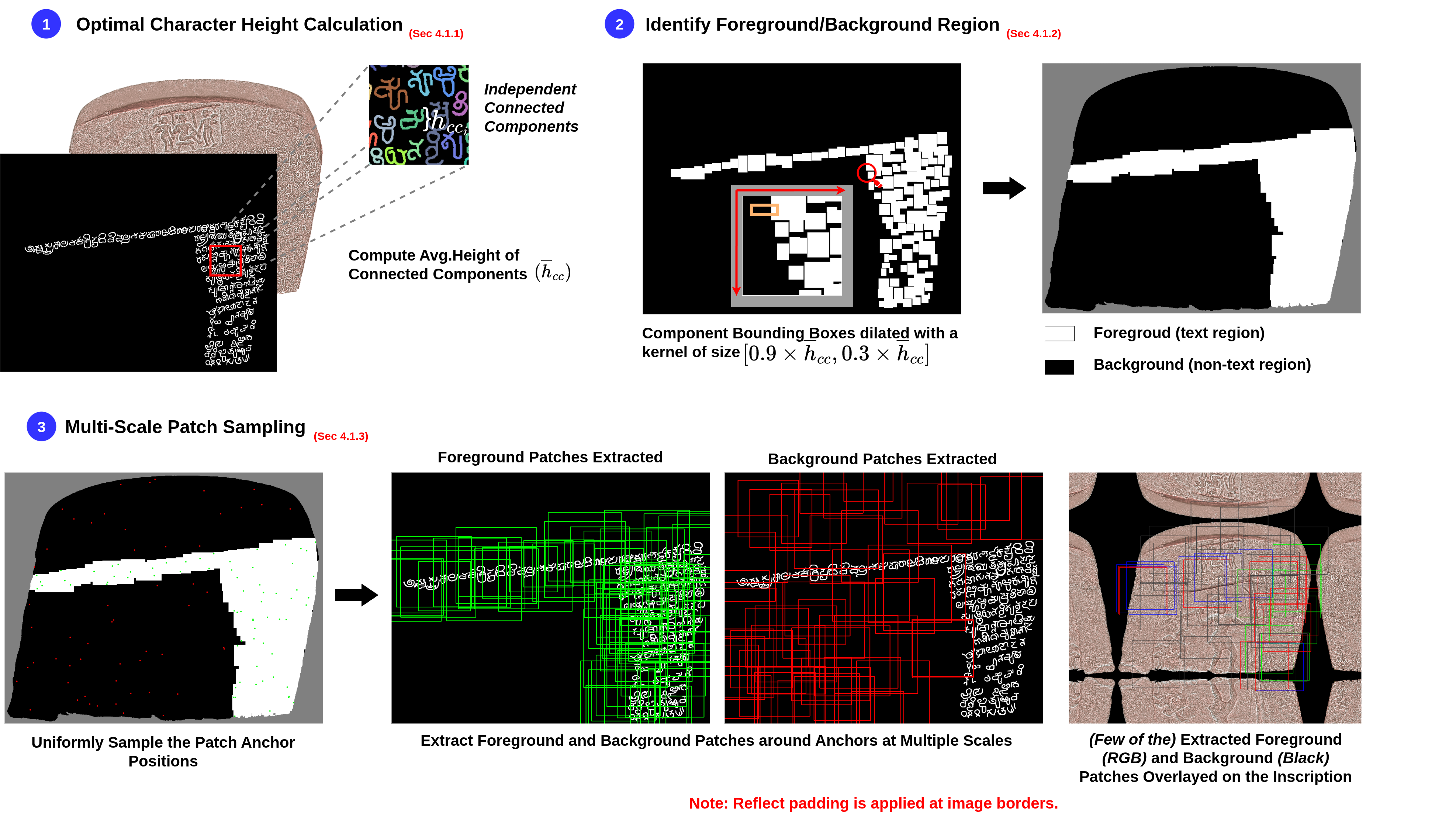}
    \caption{\textbf{Overview of our Character-Context-Aware Patch Selection Strategy}. \\ 
    \protect\bluecircled{1} First, we compute the mean character height $(\overline{h}_{cc})$ using connected components. \textit{(Sec. ~\ref{sec:proposed_approach_step_1})}\\ 
    \protect\bluecircled{2} Next, dilate (with a kernel adaptive to $(\overline{h}_{cc})$) to identify \textit{foreground (text)} and \textit{background (non-text)} regions. \textit{(Sec. ~\ref{sec:proposed_approach_step_2})} \\ 
    \protect\bluecircled{3} Finally, uniformly sample anchor points from these regions to extract multi-scale patches.  \textit{(Sec. ~\ref{sec:proposed_approach_step_3})}\\ 
    \normalfont \textit{This strategy ensures that each patch contains consistently-scaled context, enabling the model to effectively learn the distinction between character strokes and background noise.}}
    \label{fig:patching_diagram}
\end{figure*}

The inscription images used in this work were obtained from the Akshara Bhandara digital archive hosted by the Mythic Society~\cite{AksharaBhandara2024}. All images are part of the Wikimedia Commons~\cite{WikimediaCommons} public domain collection under appropriate licensing. We curated a subset of these images for annotation, prioritizing diversity in layout, surface quality, and script styles.

Annotations were created using \textit{GIMP}~\cite{GIMP} in overlay mode with a layered brush-based approach. Annotators used freehand brush tools, aided by XP-Pen drawing tablets~\cite{XPPen} for precise stroke control. Annotators were instructed to trace the centerline of strokes and fill the character body. This setup allowed accurate delineation of even shallow or partial etchings under low contrast. Depending on complexity, annotation time ranged from 15 to 20 minutes for clean, low-density images to over 2 hours for eroded, dense inscriptions. The dataset was annotated over a 1 month period by a team of 4 annotators.


The dataset is divided into training (85\%), and test (15\%) splits, stratified to reflect diversity in etching quality, surface texture, and text layout. Some salient statistics of our dataset can be seen in Table~\ref{tab:dataset-stats}. 

\begin{table}[ht]
\centering
\caption{Dataset Statistics (203  stone inscription images)}
\begin{tabular}{lcc}
\toprule
\textbf{Statistic} & \textbf{Min} & \textbf{Max} \\
\midrule
Character fragments per image & 1   & 708 \\
Image width (pixels)          & 351  & 3840 \\
Image height (pixels)         & 148  & 2784 \\
Aspect Ratio             & 0.34 & 13.3 \\
\bottomrule
\end{tabular}
\label{tab:dataset-stats}
\end{table}


\section{Proposed Approach}

We propose a novel, spatially adaptive, Character-Context-Aware patching mechanism (Sec. ~\ref{sec:patching_method}). The resulting patches are used to train a binarization network (Sec.~\ref{sec:attention_unet}). At test time, a self-refining inference pipeline (Sec.~\ref{sec:patch_merging_and_inference}) is used to intelligently mimic the training-time strategy, thereby enabling robust binarization.


\subsection{Character-Context-Aware Patching}
\label{sec:patching_method}



\begin{figure}[!t]
    \centering
    \includegraphics[width=0.9\linewidth]{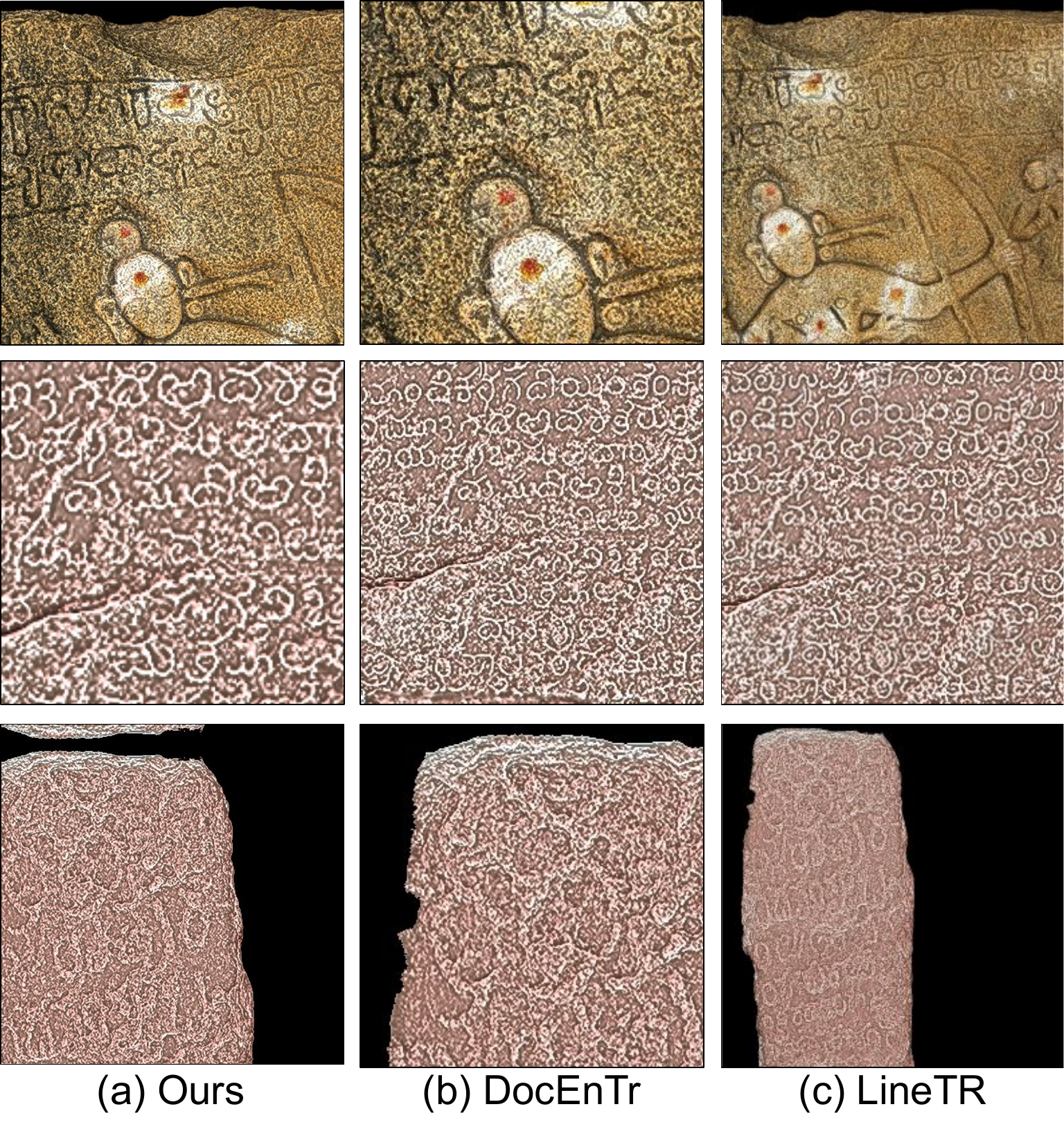}
    \caption{Our Character-Context-Aware Patching produce patches of good context, where the amount of textual information is consistent across the patches. Notice the character height is similar across the patches relative to the patch dimensions with our patching method.}
    \label{fig:patches_comparison}
\end{figure}

The foundation of our patching approach is a novel three-step strategy for generating training patches. The central idea is to use the character component itself as the fundamental unit of measurement, thereby creating a system that is inherently adaptive to the specific content of each image. Refer to Fig.~\ref{fig:patching_diagram} for blue circled items below.

\subsubsection{\textbf{Step 1: Optimal Character Height Calculation}} \protect\bluecircled{1} 
\label{sec:proposed_approach_step_1}

To make our pipeline adaptive to image content, we first calculate $\overline{h}_{cc}$ - a single, robust value representing the average character height. This step is critical since it allows all subsequent operations to be scale-invariant. We begin by identifying all connected components in the binary ground-truth mask and compute their heights $h_{\mathrm{cc},i}$ where $i$ indexes the components. To ensure that $\overline{h}_{cc}$ is not skewed by tiny components like diacritics or large non-textual elements (e.g. decorative carvings), we retain only those components whose heights fall within the inter-quartile range (IQR), i.e. 
$\mathcal{H}_{\mathrm{IQR}} = \left\{h_{cc}^{i} \;\middle|\; Q_1^h \leq h_{cc}^{i} \leq Q_3^h \right\}$ where $Q_1^{h}$ is the 25th height percentile and $Q_3^{h}$ is the 75th percentile. This statistical trimming isolates the main body of characters. $\overline{h}_{cc}$ is then computed as the mean height of this robust, filtered set, i.e. $\displaystyle \overline{h}_{cc} = \frac{1}{|\mathcal{H}_{\mathrm{IQR}}|}\sum_{h_{cc}^{i}\in\mathcal{H}_{\mathrm{IQR}}}h_{cc}^{i}$.

\subsubsection{\textbf{Step 2: Identify Foreground/Background Region}} \protect\bluecircled{2}
\label{sec:proposed_approach_step_2}

\noindent


\begin{figure}[htbp]
    \centering
    \includegraphics[width=0.9\columnwidth, trim=0 0 0 0, clip]{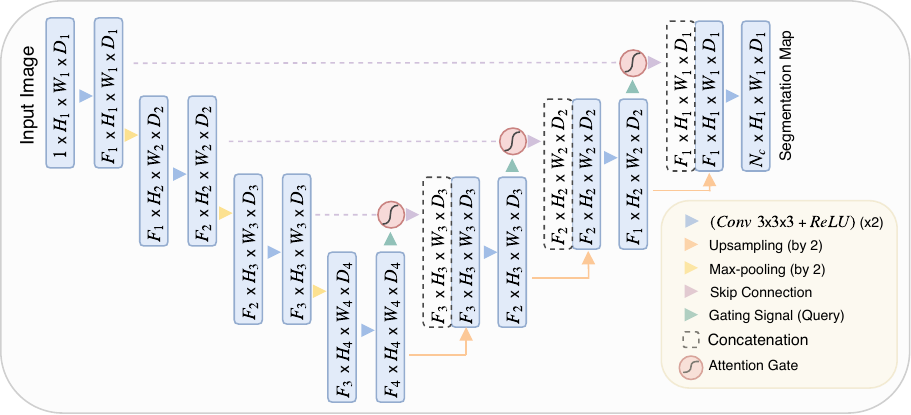}
    \caption{Architecture of the Attention U-Net used for patch-wise binarization. Attention gates modulate encoder features before concatenation at each decoder stage. \textnormal{Model architecture reproduced from \cite{Oktay2018} under CC BY 4.0 license.}}
    \label{fig:attention-unet}
\end{figure}

With $\overline{h}_{cc}$ established, we next partition the image into a foreground region (containing text) and a background region (containing only stone texture, noise, and other non-textual elements). This explicit separation allows for targeted sampling, ensuring the model is exposed to a rich and balanced set of both positive (text) and hard-negative (visually similar noise) examples. The foreground region is identified via a two-stage morphological dilation of the character components bounding boxes, where the kernel dimensions are adaptive, scaling proportionally with $\overline{h}_{cc}$. The morphological operation is done using kernel size of height of $s_1 \times \overline{h}_{cc}$ and width of $s_2 \times \overline{h}_{cc}$ for stage 1. The kernel dimensions are interchanged for stage 2. This makes the process robust to variations in character spacing and scale. It also ensures spaces between characters cannot be taken as background. The background region is the complement of this final dilated foreground area.

\subsubsection{\textbf{Step 3: Multi-Scale Patch Sampling}} \protect\bluecircled{3}
\label{sec:proposed_approach_step_3}

Finally, we extract training patches from the identified regions. To remove outlier components using a height-based criteria, we define $\mathcal{C}_{\mathrm{valid}}
  = \bigl\{\,c \in \mathcal{C}\;\bigm|\;
     Q_{1}^h - q_1\,\mathrm{IQR}_h \le h_{cc}^i \le Q_{3}^h + q_2\,\mathrm{IQR}_h
     \
   \bigr\}$. 
We then compute a preliminary foreground count $N'_\mathrm{fg} = {\lvert \mathcal{C}_{\mathrm{valid}}\rvert}{}\,R_{\mathrm{base}}$ where $R_{base}$ is the base sampling rate for the expected number of patches per valid character.
We clamp the count to obtain the final foreground count $N_{\mathrm{fg}}
  = \max\!\bigl(N_{\min},\;\min(N'_{\mathrm{fg}},\,N_{\max})\bigr)$ where $N_{\min}$ is minimum number of patches and $N_\mathrm{max}$ maximum number of foreground patches. The clamping is crucial as it prevents images with extremely high text density to dominate the training set

\begin{algorithm}[!t]
\caption{Self-Refining Inference Pipeline}
\label{alg:inference}
\begin{algorithmic}[1]
  \Require Inscription image $I$, trained model $M$
  \Ensure Final binarized map $B_{\mathrm{final}}$
  \Statex
  \State \textbf{Stage }\protect\violetcircled{1} \textbf{ : Initial Prediction (Coarse Map Generation)}
  \State $(H, W) \gets \text{size}(I)$
  \State $S \gets \{256,384,512,768\}$ \Comment{Sliding window scales}
  \State $P_\mathrm{pyr} \gets \mathbf{0}^{|S| \times H \times W}$ \Comment{Stores prediction maps for each scale in $S$}

  \ForAll{$s \in S$}
    \State $(\text{Patches},\,\text{Locs}) \gets \textsf{SlidingWindow}(I, s)$
    \State $Y \gets M(\text{Patches})$
    \State $P_\mathrm{pyr}[s] \gets \textsf{MergePatchPred}(Y,\; \text{Locs},\; H, W)$

  \EndFor
  \State $P_{\mathrm{coarse}} \gets \max_{s}(P_\mathrm{pyr}[s])$ 
    \Comment{Max-fusion over scales}
  \State $B_{\mathrm{pseudo}} \gets (P_{\mathrm{coarse}}>0.5)$
    \Comment{Generate pseudo-ground truth}
  \Statex
  \State \textbf{Stage }\protect\violetcircled{2} \textbf{: Context-Aware Refinement}
  \State $\bar{h}_{\mathrm{cc}} \gets \textsf{CalcAvgIQRHeight}(B_{\mathrm{pseudo}})$
  \State $(\text{Patches}',\,\text{Locs}') \gets \textsf{ContextAwarePatch}(I,\,B_{\mathrm{pseudo}},\,\bar{h}_{\mathrm{cc}})$
  \State $A \gets \mathbf{0}^{H \times W}$ \Comment{Accumulator for binary logits}
  \State $C \gets \mathbf{0}^{H \times W}$ \Comment{Accumulator for counts}
  \State $Y' \gets M(\text{Patches}')$
  \ForAll{each $(y,\,\ell)$ in zip$(Y',\,\text{Locs}')$}
    \State $A[\ell] \mathrel{+}= y$
    \State $C[\ell] \mathrel{+}= 1$
  \EndFor
  \State $P_{\mathrm{final}} \gets A \;\oslash\; (C + \varepsilon)$
    \Comment{Average overlapping predictions}
  \State $B_{\mathrm{final}} \gets (P_{\mathrm{final}}>0.5)$
  \State \Return $B_{\mathrm{final}}$
\end{algorithmic}
\end{algorithm}

We set a background patch count proportional to the available background area:
  $N_{\mathrm{bg}}
  = \frac{A_{\mathrm{bg}}}{A_{\mathrm{total}}}\,N_{\mathrm{bg}^{\max}}$.
The $ N_{\mathrm{bg}^{\max}}$ enforces a max limit which prevents oversampling from a text sparse images. As the background regions are generally homogeneous and less information dense than text, even sparser sampling captures enough variability. This allows us to focus the computational and learning capacity for the textual region.

With the patch counts determined, the side length of each extracted patch, \(L_{\mathrm{patch}}\), is adaptively sized based on the mean character height \(\bar{h}_{\mathrm{cc}}\):
$
L_{\mathrm{patch}} = k \cdot \bar{h}_{\mathrm{cc}}
$ where $k$ is sampled from a uniform distribution. This multi-scale approach serves as a powerful form of data augmentation, making the model inherently robust to variations in character scale. 
Refer Fig. \textit{~\ref{fig:patches_comparison}a} to view sample patches created by our patching method. Notice that the character scales relative to patch image length are consistent thanks to our patching method. Overall, the patches generated by our approach ensure the binarization model (next section) can be trained with consistent and representative views of both text and non-text regions.

\subsection{Attention U-Net Binarizer}
\label{sec:attention_unet}
\begin{figure}[t] 
    \centering
    \includegraphics[width=0.9\columnwidth]{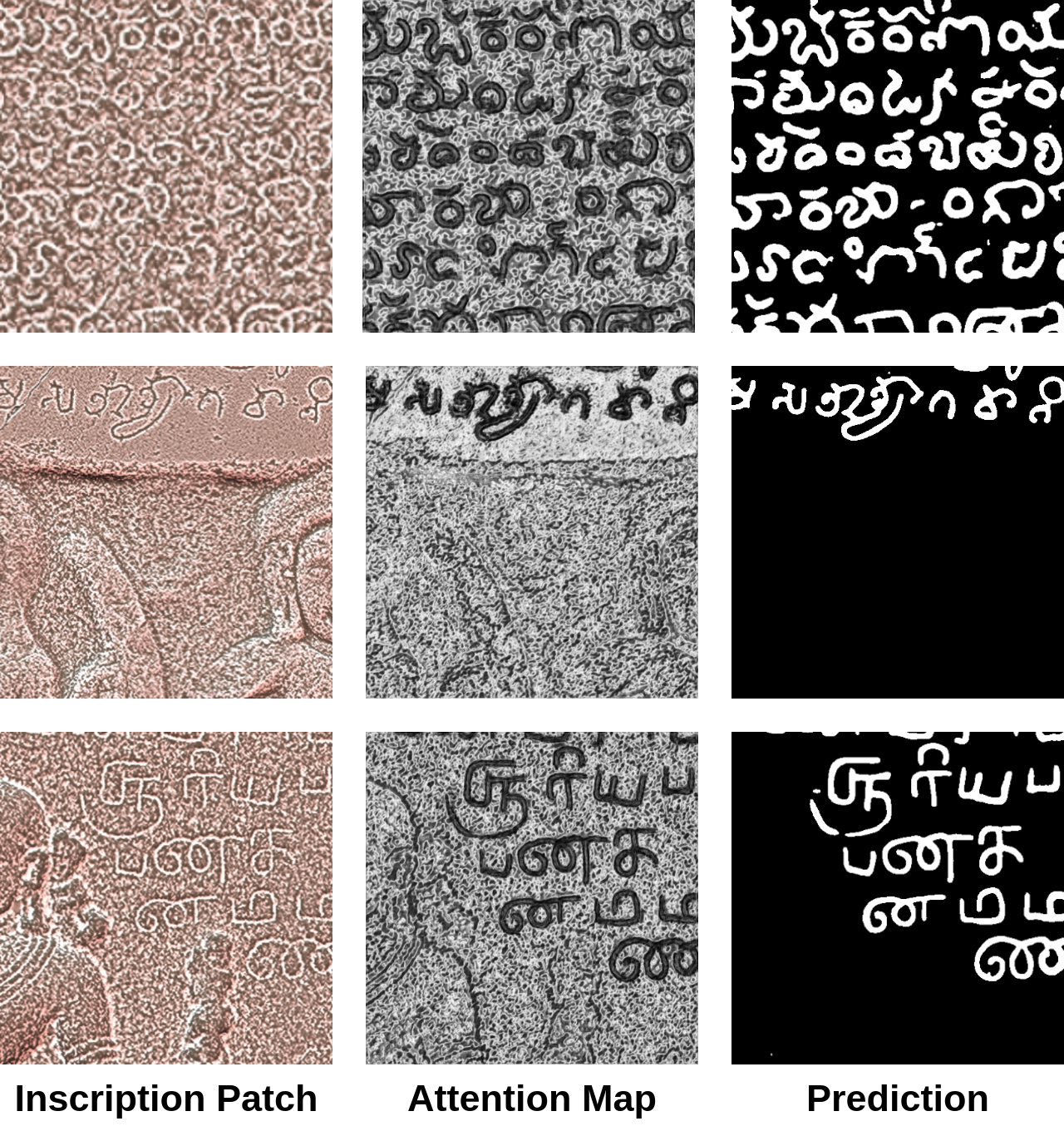}
    \caption{Attention maps extracted from the decoder layer of Attention U-Net reveal how the model learns to selectively target foreground text. Darkened regions correspond to areas the model focuses on during binarization. The model selectively enhances low-contrast strokes and suppresses background patterns.}
    \label{fig:attention-vis}
  \vspace{-10pt}  
\end{figure}

The patches obtained using our patching strategy from previous section are used to train an Attention UNet~\cite{Oktay2018} model for the binarization task. Attention U-Net extends the U-Net~\cite{Ronneberger2015} architecture by introducing attention gates at skip connections, allowing the model to focus on relevant spatial regions while suppressing irrelevant background noise. This mechanism proves especially useful in the context of inscriptions, where faint strokes and background texture are difficult to distinguish. Refer Fig.~\ref{fig:attention-unet},~\ref{fig:attention-vis}.

\begin{figure*}[t]
    \centering
    \includegraphics[width=0.9\textwidth]{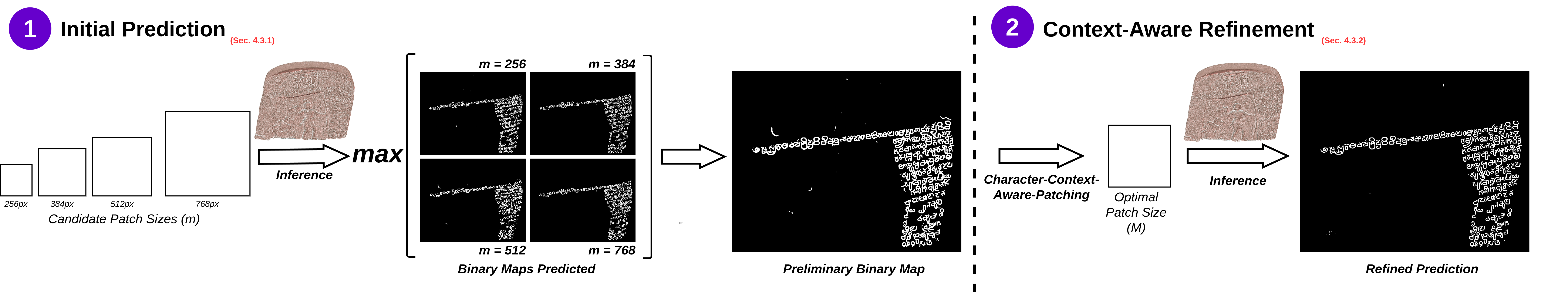}
    \caption{\textbf{Overview of our Self-Refining Inference Pipeline.} \\ 
    \protect\violetcircled{1} First, we perform inference with patch sizes \textit{256, 384, 512, 768} and fuse their predictions to get a preliminary binary map. \textit{(Sec. ~\ref{sec:inference_pipeline_step_1})}\\ 
    \protect\violetcircled{2} Using this map as a pseudo-ground truth, we then determine the optimal patch size and perform a final inference pass to get the refined prediction. \textit{(Sec. ~\ref{sec:inference_pipeline_step_2}})}
    \label{fig:inference_pipeline}
\end{figure*}

\subsection{Self-Refining Inference Pipeline}
\label{sec:patch_merging_and_inference}
During inference, our goal is to apply the learned binarization model to new, unseen stone inscription images. To ensure the patches are scaled appropriately to the character heights and to replicate the patching strategy used during training, we introduce a two-stage, self-refining inference pipeline. This pipeline first generates a coarse prediction of the text regions and then uses this information to guide a more precise, context-aware binarization. Both the stages use the same trained Attention U-Net model (Sec. ~\ref{sec:attention_unet}). Refer to Fig. ~\ref{fig:inference_pipeline} for circled numbers below.


\subsubsection{\textbf{Stage 1: Initial Prediction: }} \protect\violetcircled{1}
\label{sec:inference_pipeline_step_1}
We first perform a multi-scale prediction, where the image is processed using sliding windows of four different scales (\textit{256}, \textit{384}, \textit{512} and \textit{768} pixels). This ensures that at least one of the scale would be optimal for the characters present in the image. For each pixel, across all scales, the value (predicted from the model) with maximum probability is taken to create a preliminary binary map. This initial map, though potentially coarse, serves as a ``pseudo-ground truth" mask, providing a strong prior for identifying foreground regions in \textit{Stage 2}.

\noindent
\subsubsection{\textbf{Stage 2: Context-Aware Refinement: }} \protect\violetcircled{2} 
\label{sec:inference_pipeline_step_2}
By treating the output from \textit{Stage 1} as a binary map ground truth, we apply our proposed \textit{Character-Context-Aware Patching strategy} (Sec.~\ref{sec:patching_method}). This allows for a more targeted and dense sampling of patches from the identified foreground (text) and background regions at their optimal scales. These newly sampled patches are then passed through the trained model a second time to yield the final, refined predictions. This refinement step is crucial as it leverages patches that are optimized for character context, leading to a significant reduction in false positives and an improvement in the coherence of the binarized text boundaries (Fig. ~\ref{fig:inference_pipeline}). We employ a patch merging strategy to aggregate dense patch-level predictions into a complete binary map. Refer Algorithm ~\ref{alg:inference} for more information.

\noindent
This two-stage, self-refining process significantly enhances the final binarization quality. While the effectiveness of the second stage depends on the initial identification of text regions in the first stage, we found our multi-scale sliding window approach to be robust in practice, identifying the vast majority of foreground regions and enabling high-quality, refined predictions even in cases of hard zero-shot results, even in cases of different scripts (see Fig. ~\ref{fig:zeroshot_examples}).

\section{Implementation Details}
\label{sec:implementation_details}
\noindent\textbf{Data Preparation and Patching Hyperparameters: }
To implement our Character-Context-Aware Patching strategy (Sec.\ref{sec:patching_method}), we calibrated a set of hyperparameters that balance text-region coverage, background sampling, and multi-scale augmentation. For the kernel in Step 1 of patching, we use $s_1=0.3,s_2=0.9$. For foreground sampling, we use a base rate \( R_{\mathrm{base}} = 0.5 \), extracting $5$ patches per $10$ valid character component. In Step 2, we set IQR scaling factors $q_1=q_2=1.5$. The total number of foreground patches is clamped between $10$ ($N_{\min}$) and $250$ ($N_{\max}$). For background (negative) sampling, the limit is 75 patches per image ($N_\mathrm{bg}^{\max}$). To introduce scale variation, each patch’s side length is set to \(L_{\mathrm{patch}} = k \cdot \bar{h}_{cc}\), with \(k \sim \mathcal{U}(4, 12)\). Each patch is resized to \(512 \times 512\) pixels before passing it to the network.

\noindent\textbf{Loss Function for Attention U-Net:} We selected a hybrid DiceBCELoss function. This choice is motivated by its suitability for highly imbalanced segmentation tasks. The Binary Cross Entropy component provides stable pixel-level gradients while the Dice term directly optimizes the F1-score (segmentation overlap), encouraging the model to produce spatially coherent and complete character shapes. The Dice loss to BCE loss are equally weighted. The model is trained with Adam optimizer, with learning rate of $1 \times 10^{-4}$.  We trained our model on a single Nvidia A6000 GPU with a batch size of 16 for 50 epochs while storing the checkpoints with the best Dice score.


\section{Experiments}

We evaluate our method on the test split of our inscription dataset using the standard document binarization metrics: Peak signal-to-noise ratio ($PSNR$), F-measure ($FM$), pseudo-F-measure ($F_{ps}$)~\cite{6305530} and Distance Reciprocal Distortion ($DRD$)~\cite{1261986}. We compare our binarization pipeline against both zero-shot and supervised baselines, and conduct ablation studies to assess the contribution of each component in our pipeline.

\noindent \textbf{Binarization Baselines:} \\  
We consider the baselines outlined below.
\begin{itemize}
    \item Otsu~\cite{otsu1979}: Global thresholding method that selects a threshold minimizing intra-class variance.
    \item Sauvola~\cite{sauvola2000}: A local adaptive thresholding algorithm.
    \item Standard U-Net~\cite{Ronneberger2015}: Encoder-decoder architecture with skip connections.
    \item FCN~\cite{Tensmeyer2017}: Fully convolutional network that replaces dense layers with up-sampling for semantic segmentation.
    \item NAF-DPM~\cite{cicchetti2024nafdpm}: Based on Diffusion Probabilistic Model, the current SOTA on DIBCO 2019.
\end{itemize}
\textbf{Patching Strategies} \\ 
These baselines are evaluated on the following patching strategies
\begin{enumerate}[label=(\alph*)]
    \item Character-Context-Aware Patching \textbf{\textit{(ours)}}: Patch sizes are scaled to the character component height.
    \item Context-Adapted Patching \textit{(LineTR~\cite{agrawal2025linetr})}: Patch sizes are scaled to the average interline gap between the text lines.
    \item Fixed-patching with 50\% overlap \textit{(DocEnTr~\cite{souibgui2022docentr})}
\end{enumerate}
For Otsu~\cite{otsu1979} and Sauvola~\cite{sauvola2000}, patching is not applied as they operate on global or local thresholding principles that do not require division of the image into smaller regions.

\begin{figure*}
    \centering
    \includegraphics[width=0.9\textwidth]{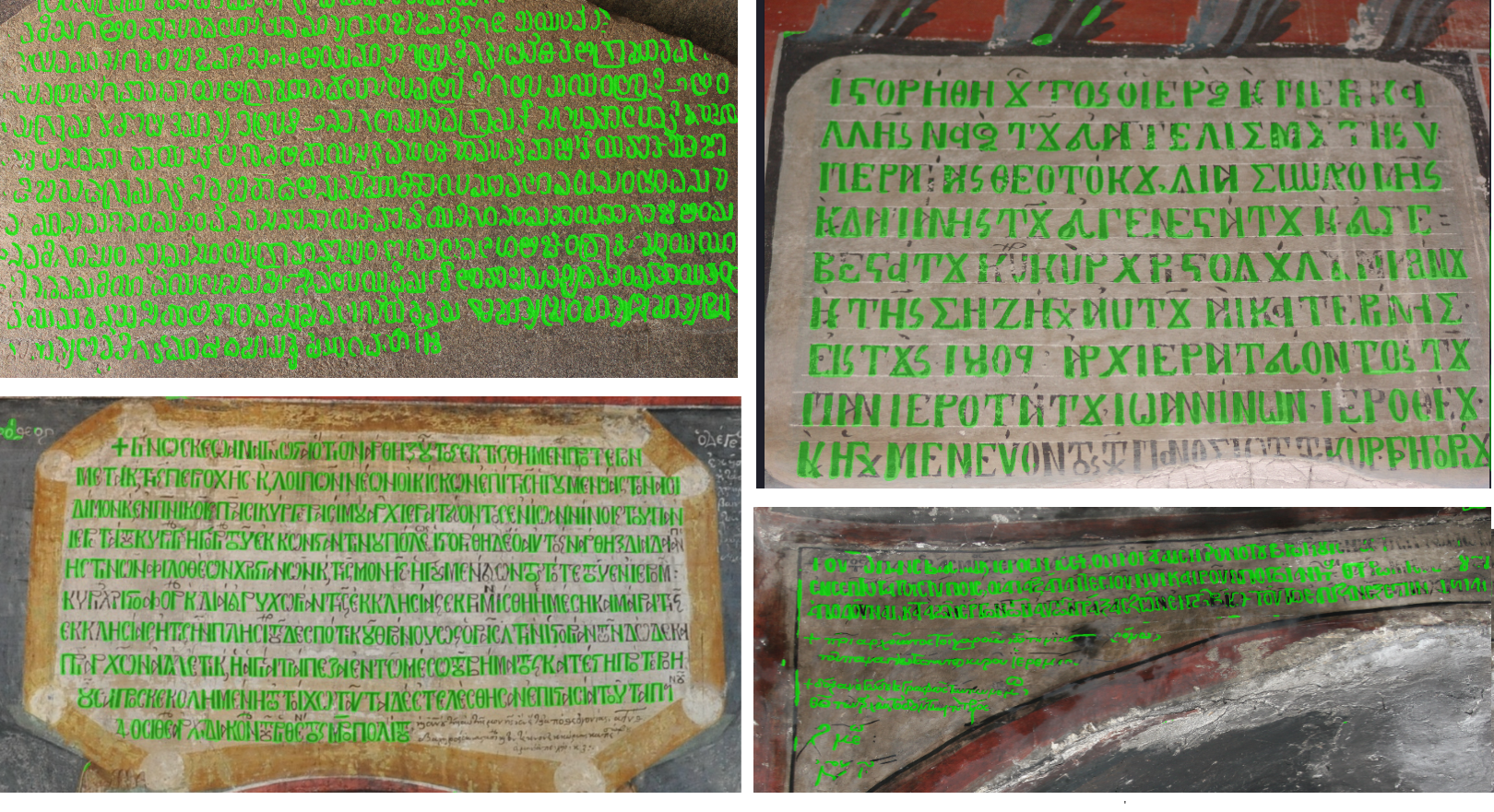}
    \caption{Demonstration of our model's robust zero-shot generalization. Examples are from challenging, in-the-wild Indic and Byzantine-era Medieval Greek~\cite{sfikas2024bessarion} inscriptions. Despite significant variations in script, lighting, and surface degradation, our method consistently produces clean, legible binarizations. \normalfont Note: The predicted binary maps are overlaid on the inscriptions.}
    \label{fig:zeroshot_examples}
\end{figure*}

\section{Results}
\label{sec:results}
\begin{figure*}[htb] 
    \centering
    \includegraphics[width=0.9\textwidth]{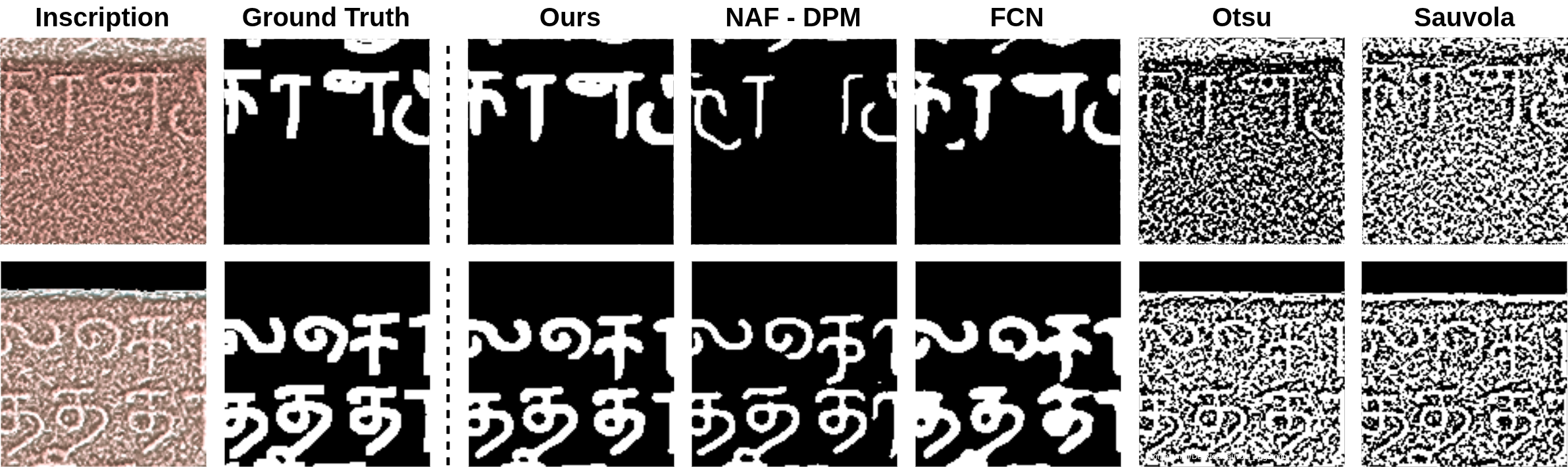}
    \caption{Qualitative Comparison between our method and other approaches. From left to right, the input inscription image and ground truth mask, and the predictions by Otsu~\cite{otsu1979}, Savoula~\cite{sauvola2000}, FCN~\cite{Tensmeyer2017}, NAF-DPM~\cite{cicchetti2024nafdpm} and our model. The characters restored by our network are clearly more readable and accurate.}
    \label{fig:inference_comparision}
\end{figure*}

Table~\ref{tab:quantitative_results} presents a comparative analysis of all methods. Our method outperforms all baselines by a significant margin, both in the test set and in zero-shot setting. Our patching strategy performs the best when coupled with Attention UNet~\cite{Ronneberger2015}. We can also observe a performance improvement when our patching strategy is coupled with other models as well. See Fig. ~\ref{fig:inference_comparision} for qualitative comparison of binary map predictions.

\begin{table*}[t]
\centering

\renewcommand{\arraystretch}{1.0}
\setlength{\aboverulesep}{0pt}
\setlength{\belowrulesep}{0pt}
\setlength{\heavyrulewidth}{0.08em} 
\setlength{\lightrulewidth}{0.05em} 

\caption{Comparison of models and patching strategies on Test Set and Zero-shot dataset.}
\label{tab:quantitative_results}
\sisetup{detect-weight=true, detect-family=true} 
\resizebox{\textwidth}{!}{%
\begin{tabular}{ll S[table-format=2.2] S[table-format=2.2] S[table-format=2.2] S[table-format=3.2] S[table-format=2.2] S[table-format=2.2] S[table-format=2.2] S[table-format=3.2]}
\toprule
\multirow{2}{*}{\textbf{Model}} & \multirow{2}{*}{\textbf{Patching}} & \multicolumn{4}{c}{\textbf{Test Set}} & \multicolumn{4}{c}{\textbf{Zero Shot}} \\
\cmidrule(lr){3-6} \cmidrule(lr){7-10}
& & {\textbf{PSNR} $\uparrow$} & {\textbf{FM} $\uparrow$} & {\textbf{F\textsubscript{ps}} $\uparrow$} & {\textbf{DRD} $\downarrow$} & {\textbf{PSNR} $\uparrow$} & {\textbf{FM} $\uparrow$} & {\textbf{F\textsubscript{ps}} $\uparrow$} & {\textbf{DRD} $\downarrow$} \\
\midrule
\multirow{1}{*}{Otsu~\cite{otsu1979}}
 & No Patching                        & 3.12 & 7.48 & 6.78 & 354.15             & 4.77 & 33.38 & 34.54 & 102.59 \\
\cmidrule(l){2-10}
\multirow{1}{*}{Savoula~\cite{sauvola2000}}
 & No Patching                        & 4.47 & 12.41 & 11.16 & 198.25           & 6.98 & 28.11 & 30.67 & 51.65 \\
\cmidrule(l){2-10}
\multirow{3}{*}{Standard U-Net~\cite{Ronneberger2015}}
 & LineTR~\cite{agrawal2025linetr}    & 13.45 & 48.27 & 53.34 & 19.20           & 8.63 & 9.52 & 10.76 & 27.30 \\
 & DocEnTr~\cite{souibgui2022docentr} & 14.55 & 59.68 & 65.85 & 13.48           & 8.77 & 19.78 & 22.06 & 25.91 \\
 & \textit{ours}                      & \bfseries 14.71 & 64.15 & 70.53 & 12.37 & 8.68 & 29.67 & 31.50 & 25.71 \\
\cmidrule(l){2-10}
\multirow{3}{*}{FCN~\cite{Tensmeyer2017}}
 & LineTR~\cite{agrawal2025linetr}    & 12.64 & 38.41 & 41.47 & 24.62           & 8.41 & 11.24 & 11.79 & 28.63 \\
 & DocEnTr~\cite{souibgui2022docentr} & 13.20 & 52.41 & 56.43 & 20.43           & 8.44 & 13.25 & 14.30 & 28.20 \\
 & \textit{ours}                      & 14.01 & 59.86 & 64.50 & 15.67           & 8.57 & 38.22 & 40.43 & 26.11 \\
\cmidrule(l){2-10}
\multirow{3}{*}{NAF-DPM~\cite{cicchetti2024nafdpm}}
 & LineTR~\cite{agrawal2025linetr}    & 13.77 & 39.42 & 45.85 & 17.96           & 8.54 & 1.87 & 2.09 & 38.56 \\
 & DocEnTr~\cite{souibgui2022docentr} & 14.28 & 51.48 & 60.26 & 15.75           & 8.74 & 16.40 & 18.93 & 37.31 \\
 & \textit{ours}                   & 14.10 & 51.98 & 61.07 & 16.31          & \textbf{9.08} & 27.66 & 30.98 & 34.30 \\
\cmidrule(l){2-10}
\multirow{3}{*}{\textbf{Attention U-Net}~\cite{Oktay2018}}
 & LineTR~\cite{agrawal2025linetr}    & 13.50 & 48.56 & 53.30 & 18.79           & 8.62 & 9.62 & 10.82 & 27.40 \\
 & DocEnTr~\cite{souibgui2022docentr} & 14.41 & 59.68 & 66.68 & 13.84           & 8.72 & 15.84 & 17.83 & 26.41 \\
 & \bfseries\textit{ours}             & 14.61 & \bfseries 66.03 & \bfseries 72.20 & \bfseries 12.14             & 8.92 & \bfseries 39.68 & \bfseries 42.30 & \bfseries 23.59 \\
\bottomrule
\end{tabular}
} 
\end{table*}

\subsection{Ablation Studies}
\label{sec:ablation_studies}
We conduct ablation studies to isolate the contribution of each major component in our pipeline (see Table~\ref{tab:ablations}).


\begin{table}[t!]
\centering
\caption{Ablation study of model variants}
\label{tab:ablations}
\begin{tabular}{lcccc}
\toprule
\textbf{Variant} & \textbf{PSNR} $\uparrow$ & \textbf{FM} $\uparrow$ & $\mathbf{F_{ps}}$ $\uparrow$ & \textbf{DRD} $\downarrow$ \\
\midrule
No Attention Gates & 14.71 & 64.15 & 70.53 & 12.37 \\
Without Patching & 12.03 & 39.69 & 41.94 & 28.43 \\
\textbf{Ours (full pipeline)} & 14.61 & 66.03 & 72.20 & 12.14 \\
\bottomrule
\end{tabular}
\end{table}

   \noindent \textit{Attention gates in U-Net Architecture: } Removing the attention gates leads to a noticeable drop in performance. Without attention, the model struggles to focus on relevant spatial regions. It becomes more easily confused by the complex textures of the stone, leading to incomplete character strokes and a higher rate of false positives in the output. 
    
    \noindent \textit{Character-Context-Aware Patching Strategy:} Eliminating our \textit{Character-Context-Aware patching strategy} significantly affects the model’s ability to distinguish characters from background artifacts. The F-Measure plummets from 72.20 to 41.94, and the DRD score worsens from 12.14 to 28.43 (see Table \ref{tab:ablations}). This is the most significant performance drop in our ablations, validating the core premise of our paper.

    \noindent \textit{Patch Size Multiplier:} As shown in Table ~\ref{tab:ablation_multiplier}, smaller size multipliers resulted in patches that are too tightly cropped, depriving the model of the surrounding stone texture needed to distinguish faint strokes from noise and leading to fragmented predictions. Conversely, larger multipliers causes the character to become too small relative to the patch, diluting the learning signal and offering no additional performance benefits.

    \noindent \textit{Multi-scale Patch Sampling:}  We evaluated whether using a range of patch scales leads to more robust models than training with a single fixed scale. First, we trained models with fixed patch size multipliers \(k\) to find the best-performing value (Table~\ref{tab:ablation_multiplier}). Performance dropped for small (\(k < 4\)) and large (\(k > 12\)) scales. Next, we compared this fixed-scale model to our multi-scale strategy, where \(k\) is sampled uniformly from \([4, 12]\). As shown in Table~\ref{tab:ablation_multiscale}, multi-scale approach outperforms the fixed baseline across all metrics, confirming that scale variation acts as strong data augmentation. By training across zoom levels and contexts, the model becomes more robust to the wide variability in real-world inscriptions.

\begin{table}[t!]
\centering
\caption{Ablation study on the fixed patch size multiplier ($k$). Each model is trained with a single, fixed patch size of $k \times \overline{h}_{cc}$. This study validates our choice of the optimal multiplier range.}
\label{tab:ablation_multiplier}
\begin{tabular}{lcccc}
\toprule
\textbf{Patch Size Multiplier ($k$)} & \textbf{PSNR} $\uparrow$ & \textbf{FM} $\uparrow$ & $\mathbf{F_{ps}}$ $\uparrow$ & \textbf{DRD} $\downarrow$ \\
\midrule
1.5  & 12.43           & 18.46           & 21.22          & 26.79           \\
3.0  & 13.33           & 53.08           & 58.12          & 19.26           \\
4.0  & 13.52           & 53.48           & 58.49          & 18.64           \\
5.0  & 12.91           & 52.02           & 55.74          & 22.6            \\
7.0  & 13.98           & 57.94           & 62.65          & 16.16           \\
9.0  & \textbf{14.2}   & 60.17           & 65.79          & 14.89           \\
12.0 & 14.17           & 60.96           & \textbf{66}    & \textbf{14.69}  \\
15.0 & 13.98           & \textbf{61.32}  & 65.93          & 15.5            \\
\bottomrule
\end{tabular}
\end{table}


\begin{table}[t!]
\centering
\caption{Ablation study comparing our multi-scale patch sampling against the best-performing fixed-scale strategy. This demonstrates the benefit of training on patches of varying sizes.}
\label{tab:ablation_multiscale}
\begin{tabular}{lcccc}
\toprule
\textbf{Patching Strategy} & \textbf{PSNR} $\uparrow$ & \textbf{FM} $\uparrow$ & $\mathbf{F_{ps}}$ $\uparrow$ & \textbf{DRD} $\downarrow$ \\
\midrule
Fixed-Scale (best, $k=12$)  & 14.17           & 60.96          & 66  & 14.69\\
\textbf{Multi-Scale ($4 \leq k \leq 12$)}      & \textbf{14.61} & \textbf{66.03} & \textbf{72.20} & \textbf{12.14} \\
\bottomrule
\end{tabular}
\end{table}



\subsection{Zero-Shot Generalization to Other Indic Scripts}

To assess the generalizability of our approach, we evaluate our model on Indic and non-Indic stone inscriptions, which were not seen during training. These images were captured in the wild using consumer-grade cameras and manually annotated by us. Despite notable visual and structural differences from the training data, our binarization model successfully extracts foreground strokes (See Fig .~\ref{fig:zeroshot_examples}). These results suggest that our binarization model focuses on generic edge and shape cues, rather than script-specific features, demonstrating its potential for broader application in epigraphic analysis across diverse Indic scripts.

\section{Conclusion}
In this paper, we proposed a novel patching strategy and an Attention U-Net model tailored for pixel-precise binarization of challenging stone inscriptions. Our patching method generates \textit{Character-Context-Aware} patches, that capture optimal amount of textual information, enabling the model to better distinguish character regions from background artifacts. Extensive experiments demonstrate that our method significantly outperforms both traditional and modern baselines, including the current state-of-the-art on the DIBCO 2019 benchmark. We also show that by adapting our patching mechanism, the performance of existing methods can be further improved. 

Moreover, our model exhibits strong zero-shot generalization to unseen Indic and Western stone inscriptions, indicating that it learns script-agnostic structural patterns rather than language-specific features. These results highlight the potential of our approach as a robust preprocessing step for downstream tasks such as script identification, OCR, transliteration, and linguistic analysis across diverse ancient scripts. 

More broadly, our work contributes to the development of robust tools for digital epigraphy and supports the large-scale computational study of South Asian textual heritage.

\vspace{15pt}

\section{Acknowledgments}
This work is supported by Ministry of Electronics and Information Technology
(MeiTY), Government of India. We also acknowledge Akshara Bhandara for providing the dataset images used in this research. 

\bibliographystyle{ACM-Reference-Format}
\bibliography{acmart}

\end{document}